\documentclass[letterpaper, 10 pt, conference]{ieeeconf}  % Comment this line out if you need a4paper

\IEEEoverridecommandlockouts                              % This command is only needed if 
\usepackage{graphicx}
\usepackage{amsmath}
\usepackage{amssymb}
\usepackage{booktabs}
 \usepackage{wrapfig}
 \usepackage[table]{xcolor}% http://ctan.org/pkg/xcolor
\usepackage{color,graphicx,soul}

\usepackage{multirow}
\usepackage{multicol}
\usepackage{balance}

\soulregister{\cite}{7}

\title{\LARGE \bf
Multi-View Keypoints for Reliable 6D Object Pose Estimation
}

\author{Alan Li and Angela P.~Schoellig% <-this % stops a space
\thanks{The authors are with the Dynamic Systems Lab, Institute for Aerospace Studies, University of Toronto, Canada, and affiliated with the Vector Institute for Artificial Intelligence. Angela P.~Schoellig is also with the Technical University of Munich (TUM). E-mails: { \{firstname.lastname\}@robotics.utias.utoronto.ca}} 
\thanks{This work was supported by resources from Epson Canada.}% <-this % stops a space
}

\begin{document}

\maketitle
\thispagestyle{empty}
\pagestyle{empty}

%%%%%%%%%%%%%%%%%%%%%%%%%%%%%%%%%%%%%%%%%%%%%%%%%%%%%%%%%%%%%%%%%%%%%%%%%%%%%%%%
\begin{abstract}

6D Object pose estimation  is a fundamental component in robotics enabling efficient interaction with the environment. It is particularly challenging in bin-picking applications, where many objects are low-feature and reflective, and  self-occlusion between objects of the same type is common. We propose a novel multi-view approach leveraging known camera transformations from an eye-in-hand setup to combine heatmap and keypoint estimates into a probability density map over 3D space. The result is a robust approach that is  scalable  in the number of views. It relies on a confidence score composed of keypoint probabilities and point-cloud alignment error, which allows reliable rejection of false positives. We demonstrate an average pose estimation error of approximately 0.5\,mm and 2\,degrees across a variety of difficult low-feature and reflective objects in the ROBI dataset, while also surpassing the state-of-art correct detection rate, measured using the 10\% object diameter threshold on ADD error.

\end{abstract}

%%%%%%%%%%%%%%%%%%%%%%%%%%%%%%%%%%%%%%%%%%%%%%%%%%%%%%%%%%%%%%%%%%%%%%%%%%%%%%%%
\section{Introduction}

Pose estimation of surrounding objects is a crucial yet difficult task for robotic systems that are tasked to interact robustly with their environments. In bin-picking applications, for example, objects may be assorted randomly in a bin in various poses and with various levels of occlusion. In recent years, convolutional neural networks (CNNs) have demonstrated great success in the problem of object pose estimation, particularly on datasets such as LINEMOD and T-LESS \cite{TLESS}, where objects are typically rich in texture and features, resulting in near-optimal color and depth images. However, single-view object pose estimation  performs poorly when data is missing, sparse and noisy due to surfaces being shiny, glossy or transparent, or due to  occlusion and ambiguity from other objects \cite{goog}. As a result, most methods detect large amounts of false positives, which are not easily separable from correct detections. 

The goal of this work is to demonstrate accurate and reliable pose estimation  in these sub-optimal conditions by leveraging estimates from multiple viewpoints to reduce estimation error and uncertainty. We leverage a known CAD model of the objects to generate training data for our estimation network, and to perform the ICP-based refinement step.
There are three major technical contributions  that contribute to higher overall accuracy and reliability compared to other work in the literature:

\begin{enumerate}
\item We use a scalable method of combining 2D heatmaps across multiple camera viewpoints, leveraging known camera transformations to project keypoint candidates and produce a probability density over 3D space. Compared to current approaches, this method allows for a more natural representation of pose uncertainty, while also enabling scaling to a higher number of viewpoints.

\item We use the predicted 2D object mask and keypoint uncertainties to filter the depth map for pose refinement. This allows for removal of outlier points from  the background and other objects, resulting in more accurate point cloud alignment compared to the current state-of-the-art \cite{ICP}.

\item We use a novel and robust confidence score incorporating object rigidity constraints, keypoint probabilities, and point cloud alignment error to reliably estimate the accuracy of the pose, allowing for rejection of the majority of false detections.
\end{enumerate}

\begin{figure}
  \begin{center}
    \includegraphics[width=75mm]{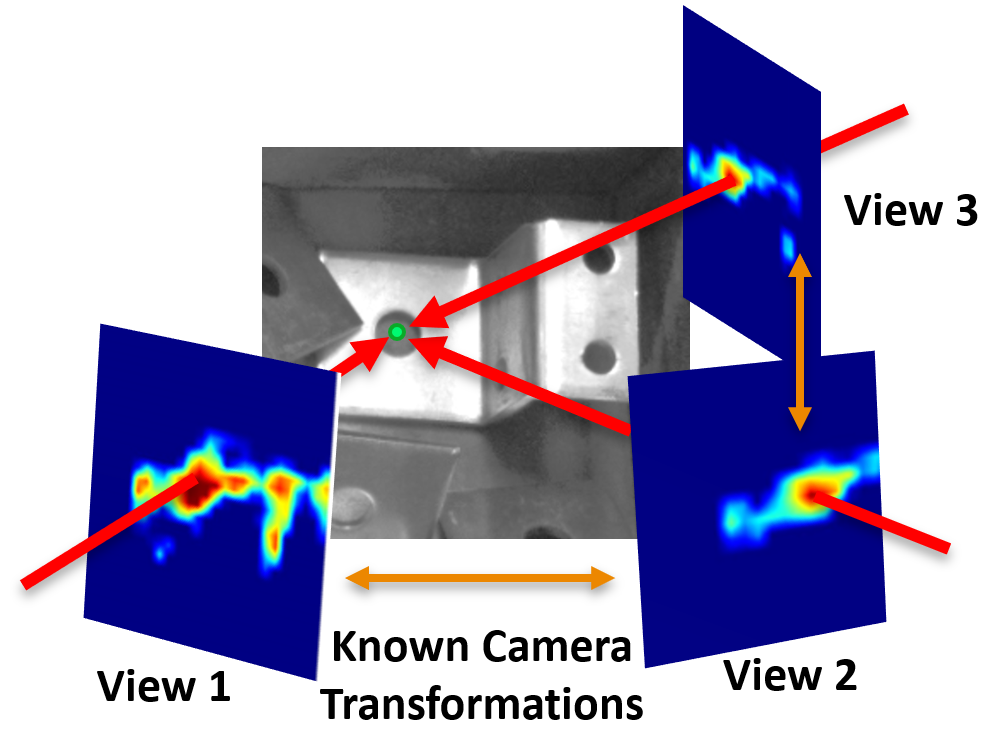}
    \end{center}
  \caption{The information from multiple viewpoints  (with known relative camera transformation) is combined to estimate an object's 6D pose. From single-view keypoint heatmaps, we calculate multi-view uncertainty estimates for filtering and ranking of candidate poses and demonstrate improved accuracy and reliability over current works.}
  \vspace{-3mm}
  \label{fig:multiview_heatmap}
\end{figure}

\section{Related Works}
\subsection{6D Pose estimation }
\begin{figure*}
  \begin{center}
    \includegraphics[width=\textwidth]{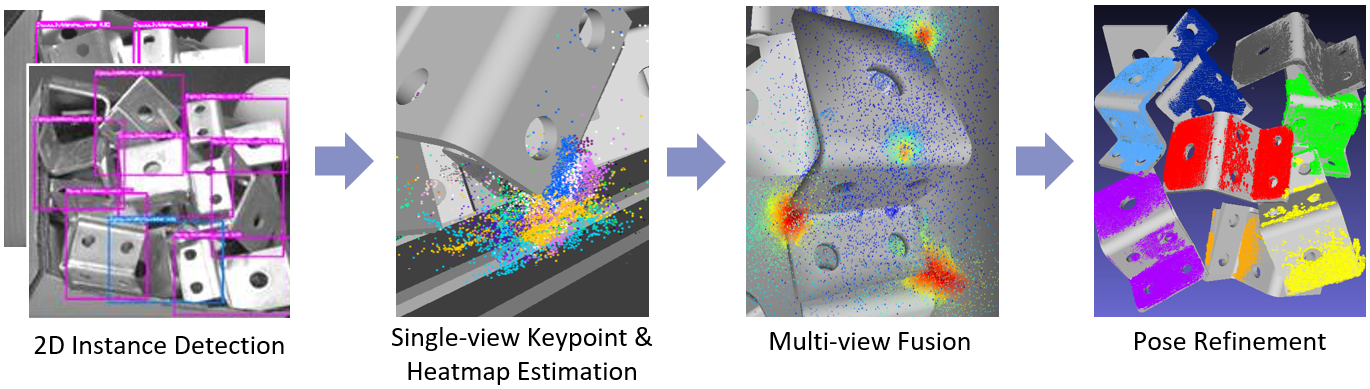}
    \end{center}
    \vspace{-4mm}
  \caption{An overview of the multi-view pipeline, showing object instance detection, followed by keypoint/heatmap computation, multi-view fusion, and finally the pose refinement step.}
  \label{fig:pipeline}
\end{figure*}
Object pose estimation from images is a core problem in computer vision, with numerous applications in robotic manipulation, augmented reality, and autonomous driving. Recent advances have leveraged the power of deep learning to surmount the shortcomings of traditional methods, which estimate pose based on local feature correspondences or template matching \cite{features}. These methods have been revisited using the advantages of CNNs to automatically learn features and implicitly estimate pose \cite{posecnn} \cite{pvnet}.

\subsection{Keypoint Representation }

The representation of 6D object pose through object keypoints has also regained popularity with the introduction of CNNs. Networks are trained to estimate the 2D or 3D location of several predefined points on an object model, which can then be used to retrieve the 6D pose of the object in the image using the correspondences between the locations of the predicted keypoints and the ones defined on the object model \cite{keypoint}. This not only simplifies the pose estimation problem into more interpretable steps, but also allows for techniques to refine and combine estimates from different viewpoints at a more fundamental level. We leverage this keypoint representation for our multi-view pipeline, allowing for more robust fusion of occluded or even partially incorrect estimates, compared to the typical 6D clustering of pose candidates of other multi-view approaches.

\subsection{Multi-view Detection }

Several recent works have explored multi-view detection using keypoints, although typically using only two or three viewpoints, often with unknown transformations between each viewpoint. A recent work combines Stereo 2D keypoint detections within a trained network to estimate the 3D positions of each keypoint for transparent and translucent objects \cite{goog}. The results show an improvement over the state of the art by a factor of 1.5-3.5, using only two known views from a stereo camera. The objects are unoccluded as opposed to placed in bins, however, and only one instance of the object can be present in the scene. While this approach is faster, the network architecture is fixed to take in exactly two stereo images as input, and to output exactly one keypoint prediction. Multi-object detection will likely prove difficult as repeated features and occlusions from different instances of the same object may confuse the depth estimation within the network. 

Another method proposes the use of multiple monocular 2D keypoint estimates from spread out viewpoints to estimate the poses of vehicles \cite{monoc}. The transformations between these viewpoints are unknown; instead the approach adds constraints based on object rigidity and the relative positions of different keypoints in the image to solve for the vehicle poses. We borrow the object rigidity contraint from this paper, but also make use of known transformations between viewpoints from a robot arm, to allow more precise fusion of estimates between views.

CosyPose is able to achieve higher performance in more cluttered scenes by  utilizing a larger number of views  \cite{cosypose}. The scene is assumed to be static across the different viewpoints, and the poses of all objects within the scene are estimated. This is then used to estimate the camera pose across different viewpoints, followed by bundle adjustment to refine the estimates and generate a globally consistent scene across all views.

In our proposed approach, the known transformation between camera viewpoints is leveraged to transform all estimates into the world frame, and fusion of estimates from different viewpoints occurs at the keypoint level, as opposed to the pose level, improving the robustness of the resulting pose estimates in the presence of occlusion.

\section{Method}

The proposed approach involves a pipeline of several steps, with the main contributions found in the multi-view fusion and pose refinement steps, shown in Fig \ref{fig:pipeline}. First, the 2D bounding boxes for each object instance in each viewpoint are detected; the cropped bounding boxes are then used to detect the keypoint locations and heatmaps. The detections are then projected into the same frame of reference and used together using the known camera poses, and finally the extracted object poses are refined using the filtered point cloud of the scene.
\subsection{2D Object Instance Detection }
In the first stage, object instances are detected in the form of 2D bounding boxes. The network is derived from YOLO object detection \cite{yolo}, specifically trained using multi-view RGB-D data that is rendered synthetically using domain randomization [7]. The network is trained to output a tight 2D bounding box centered on each object instance; the detections are then cropped and resized to 128x128 before being passed into the keypoint and heatmap detection network.

\subsection{Single-view Keypoint and Heatmap Estimation } The keypoint detection network is based off PVNet, a pixel-wise voting network for 6D pose estimation \cite{pvnet}. The network takes as input an image patch containing an object of interest, and predicts a keypoint location from each pixel input, in the form of an offset to the pixel location itself. The result is a large number of votes, one from each input pixel, of the predicted keypoint location; these votes form a cluster of keypoint candidates in 3D space. The network also predicts an object mask, which is used to filter out pixels that are not relevant to the object. Although only three keypoints are needed to determine 6D pose, we choose to instead estimate 10 to add redundancies and improve the robustness in the presence of occlusions. The main modification to the network is addition of a heatmap output, where the probability distribution of the keypoint location over the input image is estimated. This heatmap estimate is used in the next step to estimate the multi-view probabilities of keypoint candidates. The training data is generated synthetically by simulating the random arrangement of parts within a bin, followed by realistic rendering of the scene to RGB in Blender. The depth images are rendered using NxView, a depth camera simulator supplied with the Ensenso camera A total of 500 different scenes are rendered for use as training data, which are cropped around individual object instances and augmented using Domain Randomization \cite{domainrand}.

\begin{figure}
  \begin{center}
    \includegraphics[width=\linewidth]{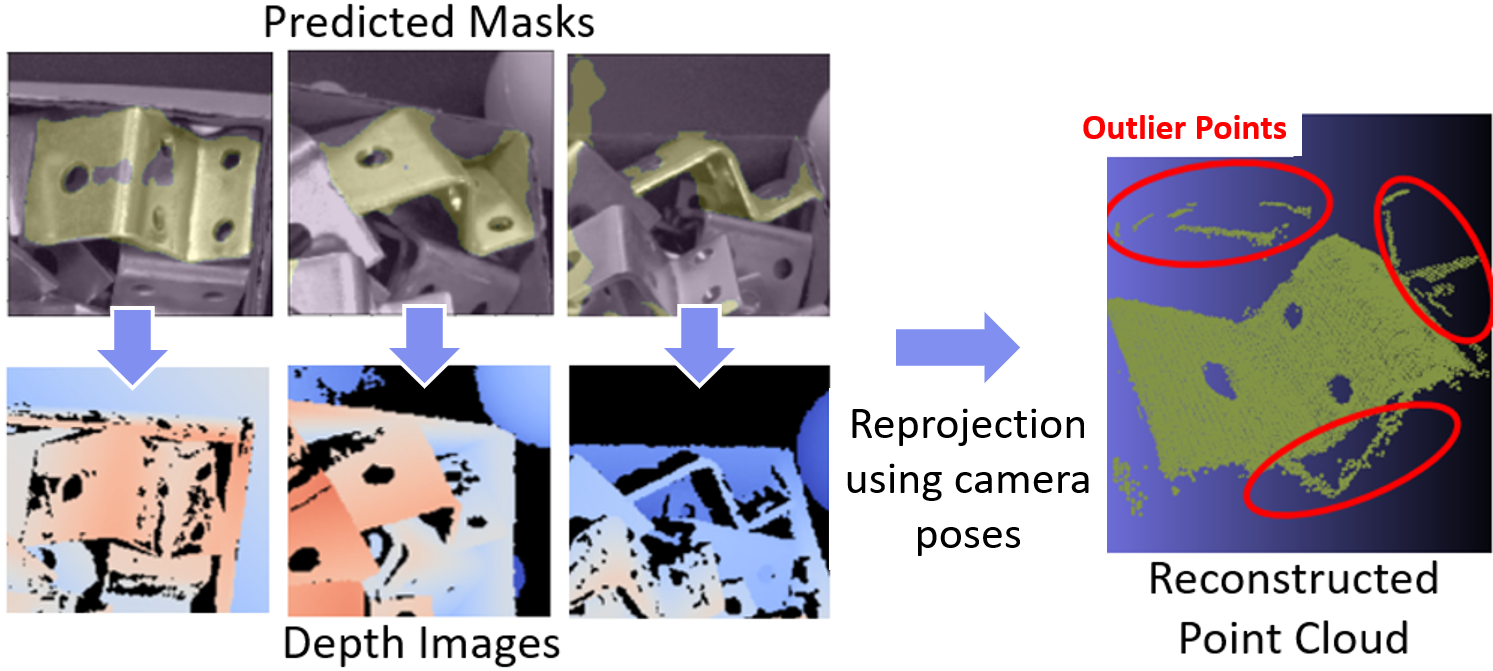}
    \vspace{-3mm}
    \end{center}
  \caption{Reconstruction of the object instance point cloud by applying the predicted mask to the depth images and projecting them to the world frame. The process results in noise and outliers around the edges due to imperfect mask prediction.}
  \label{fig:pcl_generation}
\end{figure}

\begin{figure}
  \begin{center}
    \includegraphics[width=\linewidth]{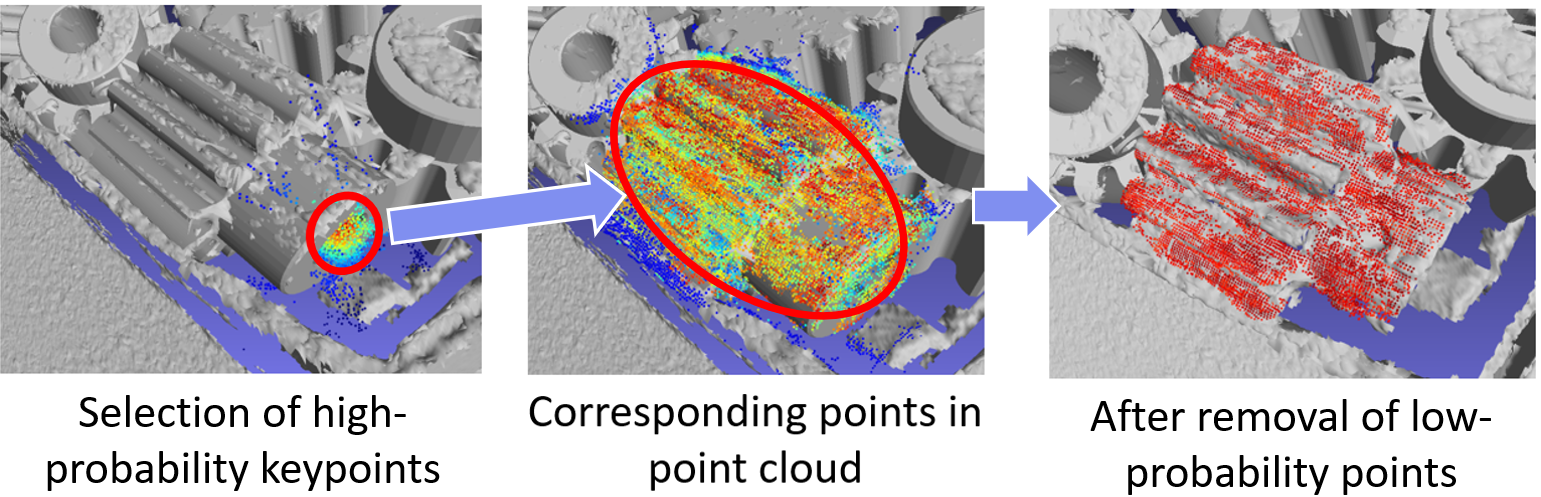}
    \vspace{-3mm}
    \end{center}
  \caption{Visualization of point cloud filtering by keypoint correspondence. By selecting only the highest probability keypoints, we can filter out outliers and noisy points in the object point cloud.}
  \label{fig:pcl_selection}
\end{figure}

\begin{figure*}
  \begin{center}
    \includegraphics[width=\textwidth]{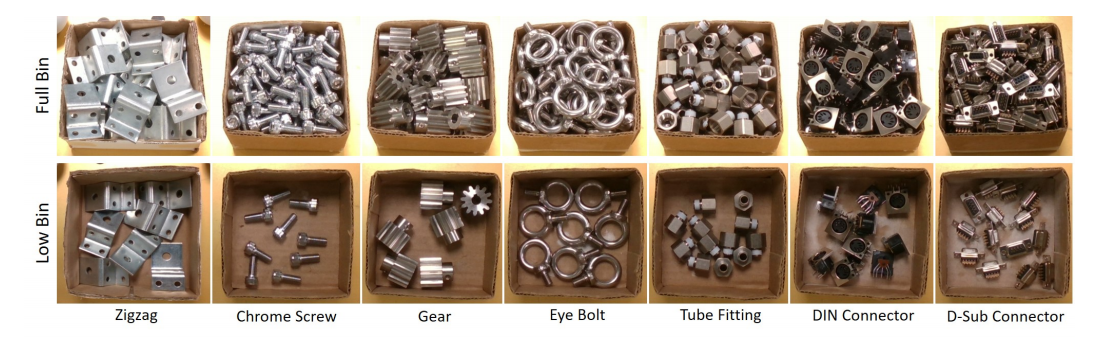}
    \vspace{-7mm}
    \end{center}
  \caption{An overview of the ROBI test dataset, which includes seven objects and two bin fill levels \cite{robi}.}
\end{figure*}

\subsection{Multiview Fusion }
After obtaining keypoint estimates from each viewpoint, the keypoints are projected into the world frame using the known camera poses of each view. This allows for multi-view clustering and object association between different views. Objects are clustered based on their center keypoint, and all keypoints within 2mm of the cluster center are determined to be the same object. Since each viewpoint also has a corresponding 2D heatmap of keypoint probabilities, we can calculate the estimated probability over all views of a given 3D keypoint candidate in world by projecting the 3D point into each of the heatmaps and retrieving the probability at the resulting pixel, as shown in Fig. ~\ref{fig:multiview_heatmap}.

The probability of a keypoint being at a point $\boldsymbol {p}_k$ in the world frame is as following:

\begin{equation}
 Pr(\boldsymbol{p}_k)=\prod_{i}^{n}h_i(u^i_{k}, v^i_{k})  ,
\end{equation}
 
where $h_i(u, v)$ represents the heatmap probability at pixel $(u, v)$ in viewpoint $i$, and $n$ represents the total number of viewpoints. The pixel coordinates $u^i_{k}$ and $v^i_{k}$ of keypoint candidate $\boldsymbol {p}_k$ in viewpoint $i$ are found by projecting the keypoint to the $i^{th}$ camera image coordinates using intrinsic matrix $\boldsymbol K$ and transformation from the world frame to the $i^{th}$ camera pose $\boldsymbol {T}^{c_i}_{w}$: 
 
\begin{equation}
\begin{bmatrix}
 u^i_{k}\\
 v^i_{k}\\
 1
 \end{bmatrix} =\boldsymbol K  \dfrac{1}{z^{c_i}_k}\boldsymbol{T}^{c_i}_{w}\boldsymbol{p}_k , 
\end{equation}
where ${z^{c_i}_k}$ is the z-coordinate of the keypoint candidate projected to the camera frame. Points which end up being projected outside of the image patch receive a probability of zero. Taking the negative log likelihood, the uncertainty of the $k^{th}$ keypoint $U(\boldsymbol{p_k})$ can be represented as:

\begin{equation}
U(\boldsymbol{p}_k) = \sum_{i}^{n}-log(h_i(u^i_{k},v^i_{k})). \
\end{equation}

The result after calculating the uncertainty for all candidates over all viewpoints is a probability density in 3D space, representing the likelihood of a given keypoint position.

The next step is the retrieval of the 6D object pose from the keypoint correspondences between the scene and the original CAD model. We use a RANSAC-based approach to sample and estimate candidate poses based on the previously calculated probability densities \cite{ransac}. The goal is to find the object pose which minimizes the overall keypoint uncertainty subject to the object rigidity constraint, which constrains the relative positions between the 10 keypoints. We define the rigidity constraint as a 6D rigid transformation matrix from the original keypoints defined on the object to the candidate object pose. 

Thus we can represent each keypoint candidate $\boldsymbol {p}_k$ as the original keypoint ${o}_k$ defined on the object transformed by the candidate object pose, represented as rotation matrix $\boldsymbol R$ and translation $\boldsymbol t$:

\begin{equation}
\boldsymbol p_k =\boldsymbol Ro_k+\boldsymbol t .\\ 
\end{equation}

The maximum likelihood estimate would then be the transformation which minimizes the keypoint uncertainties across each of the 10 keypoints $k$ in $K$:

\begin{equation}
\min\sum_{k\in K}{U(\boldsymbol p_k) =\min_{\boldsymbol R,\boldsymbol t}\sum_{k\in K}U(\boldsymbol Ro_k+\boldsymbol t)} .\\ 
\end{equation}

We approach this optimization with a heuristic method, using RANSAC to robustify against outliers. We produce candidate poses $\boldsymbol R$ and $\boldsymbol t$ by sampling three keypoints out of the ten keypoint groups according to their probability density maps, and estimating the transformation to the corresponding origin keypoints on the object model. The transformation is then used to reproject all ten keypoints on the model into the probability density maps, allowing us to calculate the overall keypoint uncertainty over all ten keypoints. The top 5 poses which result in the lowest uncertainty are selected for the next refinement step.

\subsection{Pose Refinement using Filtered ICP }

Following the poses estimated from the previous step, a refinement step is used to further align the pose estimates with the depth images obtained from the scene. Using the object mask predicted from the keypoint network, the depth images are masked such that only the object of interest remains.

These filtered depth images are then projected to into the world frame and combined using the known camera poses of each viewpoint, producing a rough 3D point cloud of each object. These point clouds contain noise and outlier points from surrounding objects and surfaces which may affect alignment quality, however, and thus a filtering step is performed first. 

Recall that for keypoint prediction, each pixel within the predicted object mask is used to estimate an offset towards the keypoint position. Additionally, each keypoint candidate is assigned an uncertainty score, based on the heatmap probability after projecting it to each viewpoint. This probability can also be related back to the original pixel within the mask which produced said keypoint prediction, giving a quality score for each point in the object point cloud. This enables the filtering of the point cloud by keypoint uncertainty, removing outliers and noisy points, as points which do not lay on the surface of the object have low probability of producing an accurate keypoint estimate.

These reconstructed, filtered 3D point clouds are then used in the Iterative Closest Point algorithm (ICP) \cite{ICP} to align the candidate poses with the observed depth data, producing the final refined outputs.

\subsection{Confidence/Uncertainty Score Calculation.}

The final confidence score given to each candidate pose is a combination of the ICP alignment error, along with the overall keypoint uncertainty. Since the camera viewpoints all tend to be within a cone above the object, the keypoint uncertainty tends to have higher variance in the Z-direction, parallel to the camera view, and lower variance, and therefore higher sensitivity in the X and Y directions. In contrast, the ICP alignment error measures mainly the Z-error, as the point cloud defines a surface of depth values spread across the X and Y axes. By combining these two metrics, we are able to create a reliable uncertainty score which can be used to safely reject the majority of false detections, while minimizing the number of correct detections being rejected.

\begin{figure*}
  \begin{center}
    \includegraphics[width=\textwidth]{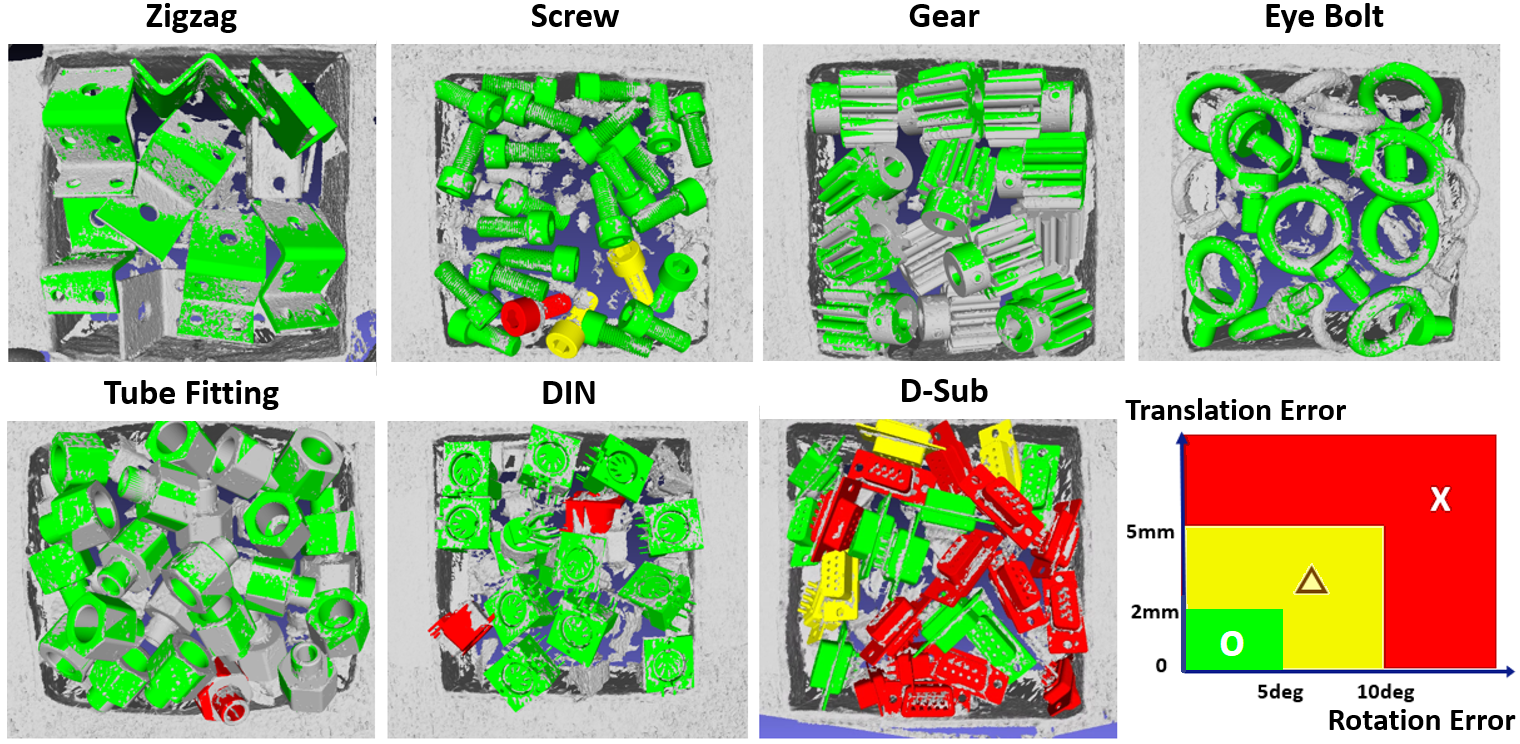}
    \end{center}
    
  \caption{Sample detections for bins of each object. Estimation performance is shown by color, with the color scheme being defined at the bottom right. Higher error cases can be seen when entire faces of the object are occluded by the bin walls or other objects. Correct detections (green) show an average accuracy of approximately 0.5\,mm and 2\,degrees on average across all objects.}
\end{figure*}

\begin{table*}[h]
\label{table_example}
\begin{center}
 \caption{Pose Estimation Accuracy  Averaged over 5 Bins}
\begin{tabular}{c||c c c c c c c c c}
\hline
\multirow{2}{2em}{Type} & \multicolumn{7}{c}{Percentage of Detections and Average Error (mm, deg)} \\
& Zigzag & Screw & Gear & Eye Bolt & Tube Fitting & DIN & D-Sub \\
%\cline{2-11}
\hline

{\cellcolor{green!75}} & 98.2 & 86.3 & 79.6 & 84.8 & 86.1 & 69.7 & 35.3 \\ 
\multirow{-2}{*}{\cellcolor{green!75}$\circ$}& (0.6, 1.2) & (0.6, 2.2) & (0.2, 1.4) & (0.5, 1.9) & (0.3, 2.07) & (0.5, 2.1) & (0.6, 2.2)\\

{\cellcolor{yellow!55}} & 1.7 & 7.7& 7.4 & 11.1 & 8.9 & 6.6 & 12.0 \\
\multirow{-2}{*}{\cellcolor{yellow!55}$\Delta$}& (2.6, 5.5) & (1.1, 7.9) & (1.6,  7.5) & (0.8, 7.7) & (0.4, 7.5) & (1.3,  8.1) & (1.2, 8.4) \\

{\cellcolor{red!75}} & 0 & 6.0 & 12.9 & 4.2 & 1 & 23.7 & 52.7  \\
\multirow{-2}{*}{\cellcolor{red!75}$\times$}& (N/A) & (3.0, 16.8) & (3.4, 14.6) & (2.9, 20.0) & (1.9, 17.1) & (3.3, 72.4) & (4.0, 129) \\

\hline

\end{tabular}

\end{center}
\end{table*}

\section{Experiments}
The effectiveness of the multi-view approach is shown using the newly released ROBI dataset. This dataset includes a total of 63 bin-picking scenes captured with an active stereo depth Ensenso N35 sensor \cite{robi}. For each scene, a view sphere totalling 88 RGB images and depth maps are captured from an Ensenso N35 stereo depth camera, and are annotated with accurate 6D poses of
visible objects and an associated visibility score. A total of seven different objects are found in the dataset, arranged randomly within bins. The objects are metallic and highly reflective, with varying levels of symmetry. This results in a dataset representing the most difficult scenes typically found in bin picking applications, with self-occlusions between multiple instances of the same object, and reflections causing missing depth data and false RGB edges.

Table 2 shows the pose estimation results on several objects in the ROBI dataset. A total of 8 different viewpoints are used as a balance between speed and accuracy. Sensitivity analysis of the keypoint accuracy to the number of views showed the majority of the performance improvements leveled off past 10 viewpoints. The ROBI evaluation consists of the top layer of objects which are less than 40\% occluded in the majority of the chosen views. The baseline performance is measured using a method proposed by Drost et al \cite{ppf}. called point-pair features, which relies on 3D point-cloud data as input. It was chosen as it ranks amongst the top methods in the BOP dataset \cite{bop}, while still benefiting from multi-view data through the fusion of depth maps. Also tested was a multi-view estimation pipeline called CosyPose \cite{cosypose}, ranking number one overall in the BOP dataset for correct detection rate. Its approach relies on single-view pose estimation in each viewpoint, followed by a pose clustering step to combine estimates into a multi-view estimate. The implementation was taken directly from the publicly available source code online. Additionally, we use another state-of-the-art object pose estimation method as a single-view baseline: 
 DC-Net \cite{dcnet}.  The correct detection rate is defined as the percentage of visible objects in the bin associated with an output pose less than 10\% ADD error, as proposed by Hinterstoisser et al \cite{hinter}. The average distance between vertices on the ground truth pose of the model and the corresponding vertices on the predicted pose is calculated, and predictions with less than 10\% average error w.r.t. the object diameter are considered correct detections.

\begin{table}[h]
\caption{Evaluation and Comparison Results on the ROBI Dataset}
\label{table_example}
\begin{center}
\begin{tabular}{c||c c c c c}
\hline
\multirow{2}{5em}{Object} & \multicolumn{5}{c}{Correct Detection Rate ($<$10\% ADD Error) }  \\
%\cline{2-11}
  & DC-Net &CosyPose & PPF & Ours  \\
\hline
Zigzag  & 30.9 & 92.4 & 86.7 & \textbf{98.5} \\ 
%\hline
Eye Bolt  & 53.5& 75.0 & 88.7 & \textbf{96.8}\\
%\hline
DIN  & 18.7& 51.9 & 61.9 & \textbf{70.7} \\
%\hline
Gear &77.6&82.1 & 76.8 & \textbf{86.0}  \\
%\hline
Tube Fit. &74.8&85.8 & \textbf{89.5} & 87.0 \\
%\hline
Screw  & 67.7& 90.3 & 78.1 & \textbf{96.2}\\
%\hline
D-Sub &10.6&4.1 & 22.9 & \textbf{31.3} \\
\hline
\end{tabular}
\end{center}
\vspace{-4mm}
\end{table}

\begin{figure*}
  \begin{center}
    \includegraphics[width=\textwidth]{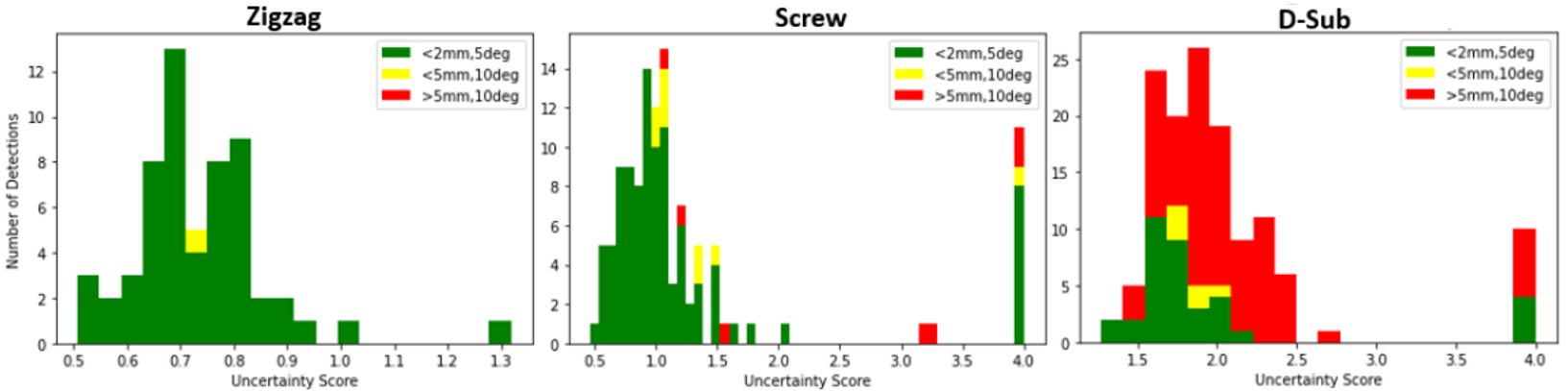}
    \end{center}
  \caption{Histogram of detection quality versus the algorithm log-uncertainty for three objects. For all objects other than D-Sub, the green detections can be separated from the majority of the yellow and red detections by thresholding based on an uncertainty score of around 1.0-1.5.}
    \label{fig:det}

\end{figure*}

\begin{figure*}
  \begin{center}
    \includegraphics[width=\textwidth]{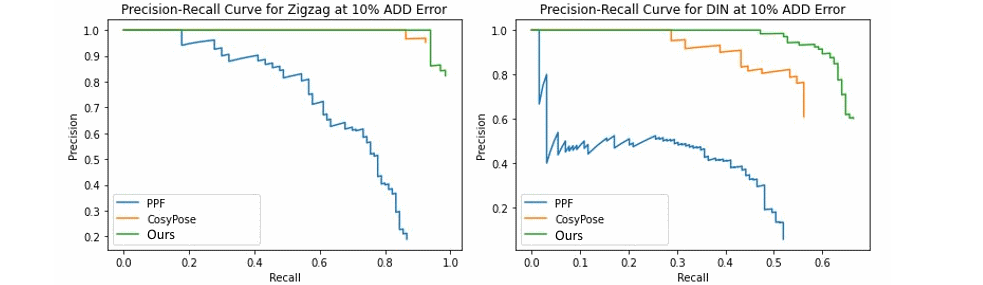}
    \end{center}
  \caption{Plots of the precision-recall curve for Zigzag and DIN objects, demonstrating the effectiveness of our confidence score in maintaining maximum precision while also achieving high detection rate.}
    \label{fig:pr}

\end{figure*}

\section{Discussion and Limitations}

The correct detection rate of our method outperforms the others in all but one object, the Tube Fitting, where PPF performs slighly better. However, looking at the precision of PPF reveals that it achieves this higher detection rate at the cost of many false positives; its confidence score is unable to sufficiently distinguish between correct and incorrect detections due to the lack of distinctive features in its depth-only approach. While CosyPose achieves higher overall precision in two objects, our method attains a larger area under the precision-recall curve, meaning the lower accuracy detections are given lower confidences and can be filtered out while minimizing the effect on correct detections, as seen in Fig.~\ref{fig:pr}. The single view DC-Net suffers particularly for Zigzag and Eye Bolt objects, which are reflective and tend to have large areas of missing depth--this is where the multi-view approaches show the largest gains. 
We found that the majority of false detections of our approach occur when the object is occluded on multiple sides, particularly by the bin walls. This results in lower accuracy, as the bin effectively blocks visibility of one side of the object, regardless of the viewpoint used. This is reflected in the uncertainty score however, which can be used to filter out the majority of these cases. The D-Sub connector proves difficult for all approaches, as the near symmetry in two axes gives four poses which all appear nearly identical from many viewpoints.

Fig.~\ref{fig:det} visualizes the separability of the three detection types, in the form of a histogram sorted by the algorithm uncertainty score. By thresholding detections based on the uncertainty, we can retain perfect precision for approximately 50\% of all correct detections in green before the higher error yellow and eventually red detections are observed. This allows us to select the confidence threshold depending on the precision required for the use case, with more detections coming at the expense of a slightly higher chance of error.
\balance

\section{Conclusion}

The main contribution of this paper is a novel method of joining keypoint estimates from different views in a natural, probabilistic way by leveraging 2D heatmaps and known camera transformations from an eye-in-hand camera, allowing for a highly-multiview approach to increase accuracy and produce reliable certainty estimates for each pose. The key advantages over existing methods include a scalable approach allowing for higher accuracies with an increasing number of views, as well as a large improvement in reliability stemming from the novel confidence score that rejects poses with higher error efficiently. 
%%%%%%%%%%%%%%%%%%%%%%%%%%%%%%%%%%%%%%%%%%%%%%%%%%%%%%%%%%%%%%%%%%%%%%%%%%%%%%%%

%%%%%%%%%%%%%%%%%%%%%%%%%%%%%%%%%%%%%%%%%%%%%%%%%%%%%%%%%%%%%%%%%%%%%%%%%%%%%%%%

%%%%%%%%%%%%%%%%%%%%%%%%%%%%%%%%%%%%%%%%%%%%%%%%%%%%%%%%%%%%%%%%%%%%%%%%%%%%%%%%

%%%%%%%%%%%%%%%%%%%%%%%%%%%%%%%%%%%%%%%%%%%%%%%%%%%%%%%%%%%%%%%%%%%%%%%%%%%%%%%%

\end{document}